\documentclass[10pt,twocolumn,letterpaper]{article}

\usepackage{times}  
\usepackage{helvet}  
\usepackage{courier}  
\usepackage{amssymb}
\usepackage{amsmath}
\usepackage{multirow}
\usepackage{float}
\usepackage{graphicx}
\usepackage{diagbox}
\usepackage{caption}
\usepackage{algorithm}
\usepackage{color}
\usepackage{algorithmic}
\usepackage{bbding}
\usepackage{booktabs}
\usepackage{gensymb}
\usepackage{makecell}
\usepackage{amsbsy, amsfonts}

\usepackage{bm}
\usepackage{graphicx}
\usepackage{multirow}
\usepackage{subfigure}
\usepackage{enumitem}

\usepackage{newfloat}
\usepackage{listings}
\usepackage{wrapfig}
\usepackage{amssymb,amsmath,amsthm}

\usepackage{graphicx}
\usepackage{amsmath}
\usepackage{amssymb}
\usepackage{booktabs}
\usepackage{multirow}
\usepackage{makecell}

\usepackage[capitalize]{cleveref}

\usepackage{cvpr}  

\crefname{section}{Sec.}{Secs.}
\Crefname{section}{Section}{Sections}
\Crefname{table}{Table}{Tables}
\crefname{table}{Tab.}{Tabs.}

\def\cvprPaperID{11} 
\def\confName{CVPR}
\def\confYear{2023}

\begin{document}

\title{Benchmarking the Robustness of Quantized Models}



\author{Yisong Xiao$^{1,2}$, Tianyuan Zhang$^2$, Shunchang Liu$^{1,2}$, Haotong Qin$^{1,2}$\\
$^1$ Shen Yuan Honors College, Beihang University \\
$^2$ State Key Lab of Software Development Environment, Beihang University\\
{\tt\small $\{$xiaoyisong, 19373397, liusc, qinhaotong$\}$@buaa.edu.cn}
}


\maketitle

\begin{abstract}
    
    Quantization has emerged as an essential technique for deploying deep neural networks (DNNs) on devices with limited resources. However, quantized models exhibit vulnerabilities when exposed to various noises in real-world applications. Despite the importance of evaluating the impact of quantization on robustness, existing research on this topic is limited and often disregards established principles of robustness evaluation, resulting in incomplete and inconclusive findings. To address this gap, we thoroughly evaluated the robustness of quantized models against various noises (adversarial attacks, natural corruptions, and systematic noises) on ImageNet. Extensive experiments demonstrate that lower-bit quantization is more resilient to adversarial attacks but is more susceptible to natural corruptions and systematic noises. Notably, our investigation reveals that impulse noise (in natural corruptions) and the nearest neighbor interpolation (in systematic noises) have the most significant impact on quantized models. Our research contributes to advancing the robust quantization of models and their deployment in real-world scenarios.

\end{abstract}

 

\vspace{-0.2in}
\section{Introduction}
\vspace{-0.05in}
\label{sec:intro}

Deep neural networks (DNNs) have demonstrated impressive performance in a broad range of applications \cite{krizhevsky2017imagenet,DBLP:conf/cvpr/ZhaoZXLP22}. However, deploying DNNs on resource-constrained devices, such as IoT devices, poses significant challenges. To address this issue, researchers have proposed various model compression techniques, including model quantization \cite{li2021mqbench,qin2023bibench} and pruning \cite{guo2021jointpruning,guo2020multi}. Among these techniques, model quantization has become a critical approach for compressing DNNs due to its ability to maintain network structure and achieve comparable performance. This technique involves mapping the network parameters from 32-bit floating-point numbers to low-bit integers, resulting in reduced memory usage and faster inference.

Despite their impressive performance, DNNs are highly susceptible to adversarial examples \cite{liu2019perceptual,goodfellow2014explaining,liu2023x,liu2022harnessing}. Adversarial examples are perturbations that are designed to be undetectable to human vision but can easily deceive DNNs, posing a significant threat to practical deep learning applications. In addition, DNNs are vulnerable to natural corruptions \cite{hendrycks2019benchmarking} such as snow and motion blur, which are common in real-world scenarios and can significantly reduce the accuracy of DNN models. Moreover, system noises resulting from the mismatch between software and hardware can also have a detrimental impact on model accuracy \cite{wang2021real}. These phenomena demonstrate that quantized networks deployed in safety-critical applications are unreliable when faced with various perturbations in the real world.

Therefore, it is critical to conduct a comprehensive evaluation of the robustness of quantized models before deploying them to identify potential weaknesses and unintended behaviors. While numerous studies have extensively investigated the robustness of floating-point networks against various attacks and metrics, research on the robustness of quantized models \cite{alizadeh2020gradient,lin2019defensive} remains inadequate. These studies lack diversity in terms of noise sources and rely solely on small datasets, leading to inconclusive findings regarding the robustness of quantized models.

We build the robustness evaluation benchmark of quantized models. Our benchmark assesses the robustness of quantized models using 3 popular quantization methods (DoReFa, PACT, and LSQ) and 4 classical architectures (ResNet18, ResNet50, RegNetX600M, and MobileNetV2). For each method, we evaluate 4 commonly used bit-widths. Our analysis includes 3 progressive adversarial attacks, 15 natural corruptions, and 14 systematic noises on the ImageNet benchmark. Our empirical results demonstrate that lower-bit quantized models display better adversarial robustness but are more susceptible to natural corruptions and systematic noises. We identify \textit{impulse noise} and \textit{the nearest neighbor interpolation} as the most harmful sources of natural corruptions and systematic noises, respectively.

\vspace{-0.11in}
\section{Related Work}
\vspace{-0.06in}
\subsection{Network Quantization} 
\vspace{-0.06in}
Network quantization compresses DNN models by reducing the number of bits required to represent each weight to save memory usage and speed up hardware inference.
A classic quantization process (\textit{quantization} and \textit{de-quantization}) can be formulated by:
\begin{equation}
Q(r)=\mathbf{Int}(r/S)-Z,\quad r^{'}=S \cdot (Q(r)+Z)
\end{equation}
where $Q$ is the quantization operator, $r$ and $r^{'}$ is real value and de-quantized real value respectively, $S$ and $Z$ denote \textit{scale} and \textit{zero-point} respectively. Given $t$ bits, the range after quantization is determined by $[-2^{t-1},2^{t-1}-1]$. 

We here provide a brief review of the commonly used quantization methods. 
One string of research designed rules to fit the quantizer to the data distribution.
For example, DoReFa-Net \cite{zhou2016dorefa} simply clips the activation to $[0,1]$ and then quantizes it, due to the observation that most activation falls into this range in many network architectures. 
Other notable work focused on learning appropriate quantization parameters during the backpropagation process.
PACT \cite{choi2018pact} clip the activation by a handcrafted parameter and optimize the clipping threshold. 
Notice that PACT has no gradient below the clip point, LSQ \cite{esser2019learned} learns the \textit{scale} alongside network parameters by estimating the gradient at each weight and activation layer.
\vspace{-0.06in}
\subsection{Adversarial Attacks} 
\vspace{-0.06in}
Adversarial examples are inputs with small perturbations that could easily mislead the DNNs \cite{goodfellow2014explaining}. 
Formally, given a DNN $f_{\Theta}$ and an input $\mathbf{x}$ with the ground truth label $\mathbf{y}$, an adversarial example $\mathbf{x}_{adv}$ satisfies
\begin{equation}
f_{\Theta}(\mathbf{x}_{adv}) \neq \mathbf{y} \quad s.t. \quad \|\mathbf{x}-\mathbf{x}_{adv}\| \leq \epsilon,
\end{equation}
where $\|\cdot\|$ is a distance metric and commonly measured by the $\ell_{p}$-norm ($p\in$\{1,2,$\infty$\}).

A long line of work has been dedicated to performing adversarial attacks \cite{goodfellow2014explaining,madry2017towards,croce2020reliable,liu2020bias,liu2019perceptual,liu2020spatiotemporal,wang2021dual}, which can be mainly divided into white-box and black-box manners based on access to the target model. For white-box attacks, adversaries have complete knowledge of the target model and can fully access it; while for black-box attacks, adversaries have limited or even without any knowledge of the target model and can not directly access it. This paper primarily employs white-box attacks to evaluate the adversarial robustness of target models, as they offer stronger attack capabilities.

\vspace{-0.05in}
\subsection{Robustness of Quantized Models}  
\vspace{-0.05in}
A number of studies have been proposed to evaluate the robustness of floating-point networks \cite{zhang2021interpreting,tang2021robustart,liu2021training,wang2022defensive}. 

However, the robustness of quantized networks has been relatively underexplored.  
Lin \etal \cite{lin2019defensive} proposed a defensive quantization method to suppress the amplification of adversarial noise during propagation by controlling the Lipschitz constant of the network during quantization. Similarly, Alizadeh \etal \cite{alizadeh2020gradient} also designed a regularization scheme to improve the robustness of the quantized model by controlling the magnitude of adversarial gradients. In this paper, we aim to thoroughly evaluate the robustness of quantized models against multiple noises for several quantization methods, architectures, and quantization bits.

\vspace{-0.13in}
\section{Evaluation Protocols}
\vspace{-0.05in}
\subsection{Evaluation Objects}
\vspace{-0.05in}
\textbf{Dataset}. 
We conduct evaluations on the large-scale ImageNet dataset with 1,000 classes, which comprises 1.2 million training images and 50,000 validation images.

\textbf{Architectures}.
We consider four architectures, including ResNet18\cite{he2016deep}, ResNet50\cite{he2016deep}, RegNetX600M \cite{radosavovic2020designing}, and MobileNetV2 \cite{sandler2018mobilenetv2}. 
(1) ResNet18 and ResNet50 are classical backbone architectures that are widely used in various computer vision tasks.
(2) RegNetX600M is an advanced architecture with group convolution discovered through model structure search.
(3) MobileNetV2 is a lightweight network designed for efficient deployment on edge devices, featuring depthwise separable convolutions.

\textbf{Quantization Methods}.
We focus on three popular quantization methods, including DoReFa \cite{zhou2016dorefa}, PACT \cite{choi2018pact}, and LSQ \cite{esser2019learned}. For the choice of quantization bits, we adopt the commonly used set in deployments (\ie, 2, 4, 6, and 8).

For each architecture, we quantize models on the ImageNet training set starting from the same floating-point model, then evaluate their robustness against perturbations generated on the ImageNet validation set.

\vspace{-0.05in}
\subsection{Robustness Evaluation Approaches} 
\vspace{-0.05in}
Quantized models are vulnerable to various perturbations in real-world scenarios. We classify these perturbations into adversarial attacks, natural corruptions, and systematic noises, following the guidelines proposed in  \cite{tang2021robustart}.

\textbf{Adversarial attacks}. To craft adversarial perturbations with progressively increasing attack capabilities, we employ FGSM-$\ell_{\infty}$, PGD-$\ell_{1}$, PGD-$\ell_{2}$, PGD-$\ell_{\infty}$ and AutoAttack-$\ell_{\infty}$. For each attack, we set three different perturbation magnitudes (small, middle, and large). 

\textbf{Natural corruptions.} To simulate natural corruptions, we utilize 15 distinct perturbation methods from the ImageNet-C benchmark \cite{hendrycks2019benchmarking} that fall into four categories: noise, blur, weather, and digital.

\textbf{Systematic noises.} Moreover, system noises are always present when models are deployed in edge devices due to changes in hardware or software. To evaluate the impact of system noises on quantized models, we utilize pre-processing operations from ImageNet-S \cite{wang2021real}, including three frequently used decoders and seven commonly used resize modes.

\vspace{-0.05in}
\subsection{Evaluation Metrics} 
\vspace{-0.05in}
\textbf{Adversarial robustness.} For specific adversarial attacks, we measure adversarial robustness ($AR$) using model accuracy, where higher $AR$ indicates a stronger model.
For the union of different attacks, we adopt the Worst-Case Adversarial Robustness ($WCAR$) to measure adversarial robustness (higher indicates a stronger model):
\begin{equation}
WCAR = 1 - P_{(\mathbf{x},\mathbf{y})\sim \mathcal{D}}{\mathrm{Any}(f(\mathcal{A}_{\epsilon,p}^{f}(\mathbf{x})) \neq \mathbf{y}) }, 
\end{equation}
where $\mathcal{A}_{\epsilon,p}$ represents the adversary, $\mathcal{D}$ is the validation set, and $\mathrm{Any}(\cdot)$ is a function that returns true if any of the adversaries attacks successfully. 

\textbf{Natural robustness.} We adopt the average accuracy of the quantized model on all corruptions (denoted as $C$) to measure natural robustness, denoted as $NR$:
\begin{equation}
NR = \mathbb{E}_{c\sim C}({P_{(\mathbf{x},\mathbf{y})\sim \mathcal{D}}{(f(c(\mathbf{x}))=\mathbf{y})}}),
\end{equation}
where $c$ denotes a corruption method. A higher value of $NR$ means better natural robustness.

\textbf{Systematic Robustness}.
We adopt the model stability on different systematic noises (denoted as $S$) to measure systematic robustness ($SR$). Specifically, we calculate the standard deviation of accuracy as $SR$:
\begin{equation}
SR = \mathbb{D}_{s\sim S}({P_{(\mathbf{x},\mathbf{y})\sim \mathcal{D}}{(f(s(\mathbf{x}))=\mathbf{y})}}),
\end{equation}
where $s$ denotes a decode or resize method. A lower value of $SR$ means better stability towards systematic noises.

\vspace{-0.05in}
\section{Empirical Results}
\vspace{-0.05in}
In this section, we present the results mainly obtained on the ResNet18 architecture.
\vspace{-0.05in}
\subsection{Clean Accuracies}
\vspace{-0.05in}

\cref{tab:clean-acc} reports the clean accuracies of quantized models. 
Most of the quantized models maintain comparable accuracy to the 32-bit ResNet18. 
However, 2-bit PACT fails to converge, resulting in a mere 2.61\% accuracy. Therefore, we exclude it from the subsequent robustness evaluations.

\begin{table}[ht]\small
\centering
\vspace{-0.13in}
\caption{Clean accuracies of the ResNet18 for all models. ``NC" denotes not converged.}
\vspace{-0.1in}
\label{tab:clean-acc}
\begin{tabular}{@{}llllll@{}}
\toprule
Model & Method & \multicolumn{1}{c}{2bit} & \multicolumn{1}{c}{4bit} & \multicolumn{1}{c}{6bit} & \multicolumn{1}{c}{8bit} \\ \midrule
\multirow{3}{*}{\makecell{ResNet18 \\ FP: 71.06}} 
                          & DoReFa & 62.31 & 70.60 & 71.15 & \pmb{71.51} \\
                  & PACT   & NC     & 70.41 & 71.17 & \pmb{71.43} \\ 
                  & LSQ    & 65.97 & 70.52 & 71.10 & \pmb{71.31} \\
                  \bottomrule
\end{tabular}%
\vspace{-0.15in}
\end{table}

\subsection{Evaluation of Adversarial Attacks}
\begin{figure}[ht]

\vspace{-0.1in}
  \centering
    \subfigure[ResNet18 quantized by DoReFa]{
\includegraphics[width=0.46\linewidth]{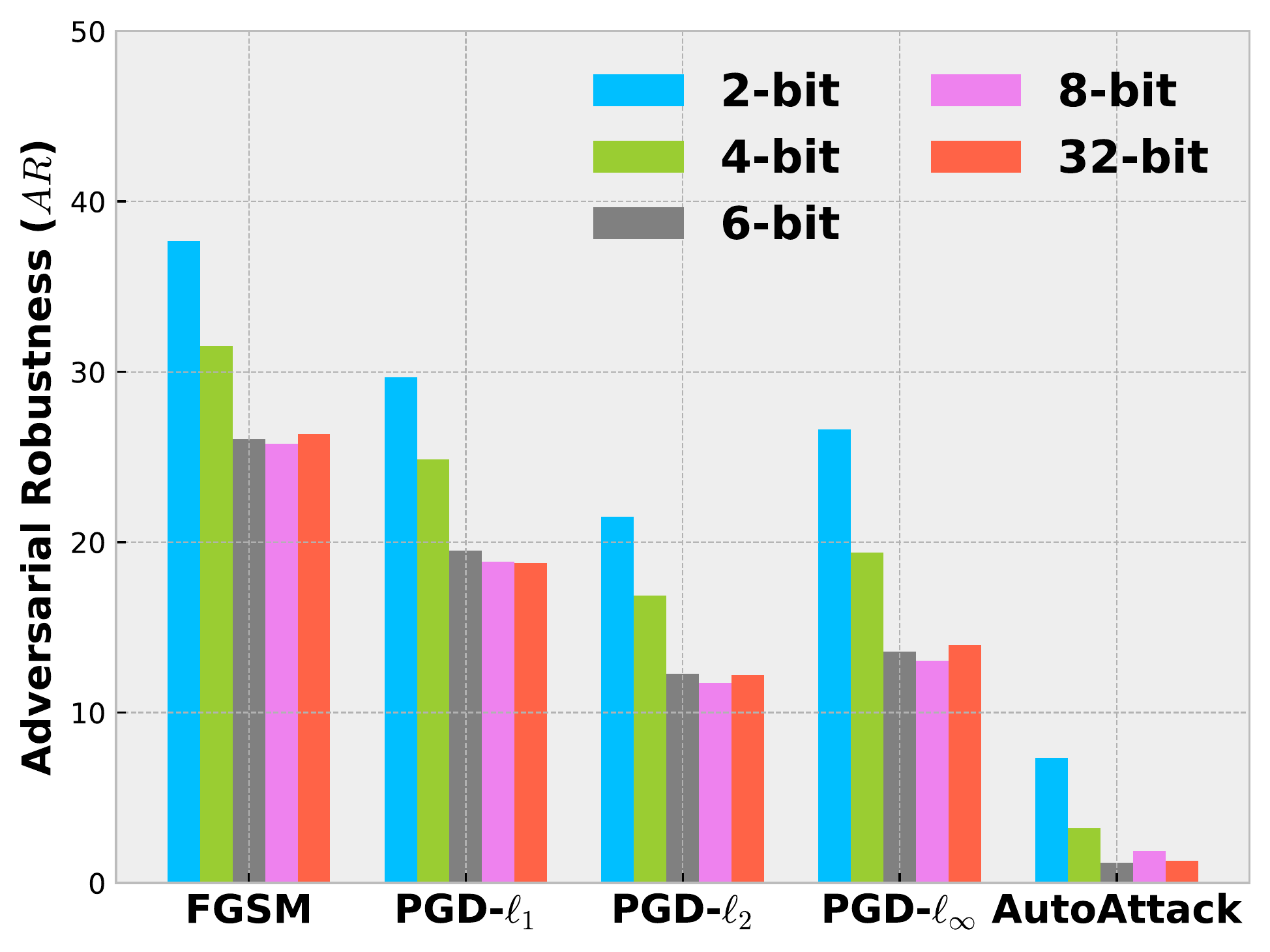}
 \label{fig:resnet18_dorefa_adv}
}
\subfigure[ResNet18 quantized by LSQ]{
\includegraphics[width=0.46\linewidth]{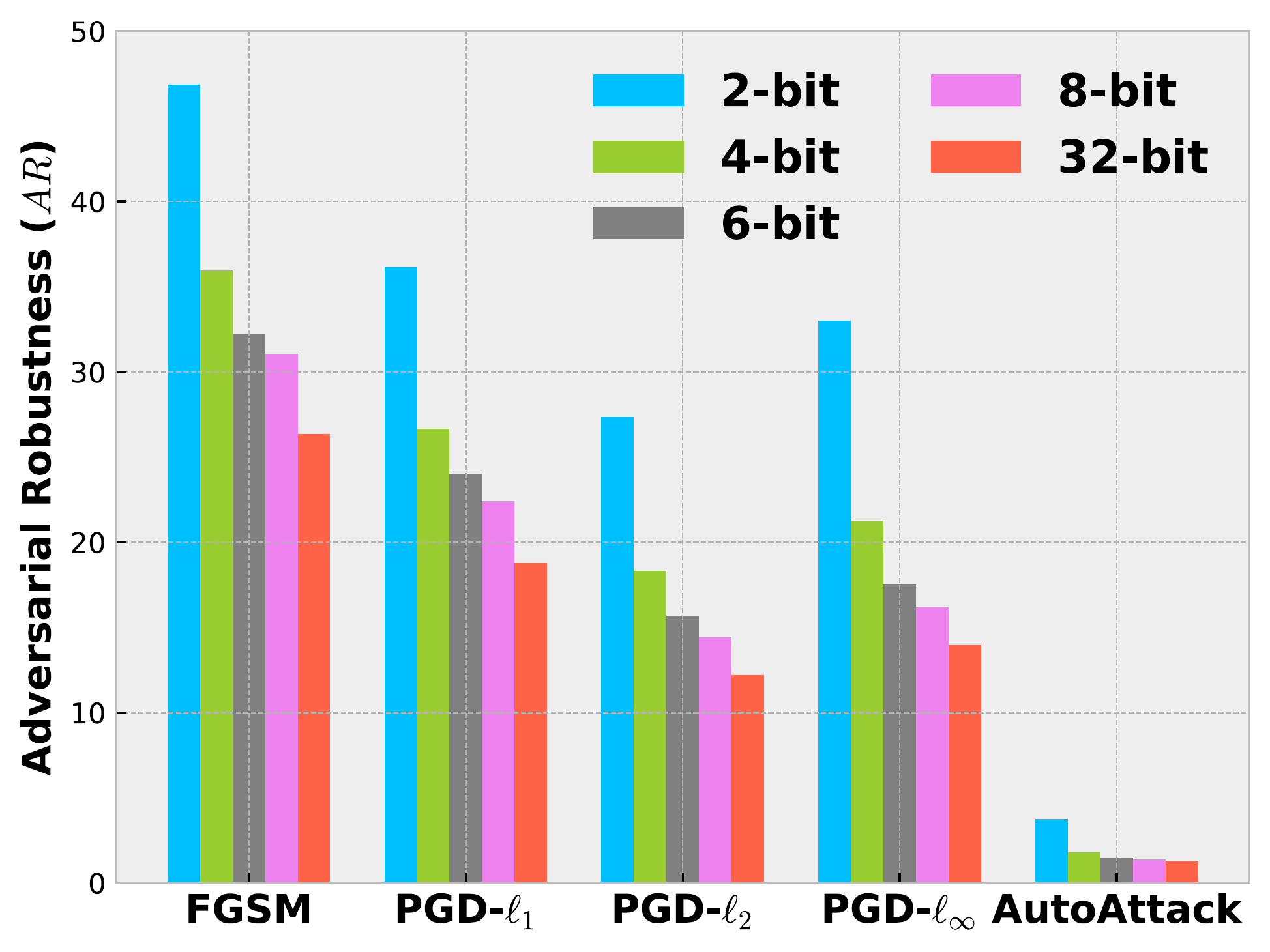}
 \label{fig:resnet18_lsq_adv}
}
\vspace{-0.1in}
  
  \caption{Adversarial Robustness against specific attacks.}
  \label{fig:adversarial-AR}
  \vspace{-0.15in}
\end{figure}

\vspace{-0.05in}
Since $WCAR$ degrades to 0 under medium and high perturbation magnitudes, we here only present the results under small magnitudes. From the results shown in \cref{tab:adversarial-WCAR}, we could make the following observations: (1) Unlike the decrease in clean accuracy, lower-bit models exhibit higher worst-case adversarial robustness and are almost better than floating-point network; (2) At the same quantization bit, PACT presents the best adversarial robustness compared to other quantization methods. Furthermore, we compare the adversarial robustness ($AR$) of quantized models against specific attacks. As depicted in \cref{fig:adversarial-AR}, we can identify that LSQ performs better under FGSM and PGD attacks, yet is vulnerable to AutoAttack (the strongest of these attacks).

\begin{table}[ht] \small
\centering
\vspace{-0.1in}
\caption{Worst-Case Adversarial Robustness ($WCAR\textcolor{red}{\uparrow}$) of ResNet18 models on small magnitude.}
\vspace{-0.1in}
\label{tab:adversarial-WCAR}
\begin{tabular}{@{}llllll@{}}
\toprule
Model & Method & \multicolumn{1}{c}{2bit} & \multicolumn{1}{c}{4bit} & \multicolumn{1}{c}{6bit} & \multicolumn{1}{c}{8bit} \\ \midrule
\multirow{3}{*}{\makecell{ResNet18 \\ FP: 1.30}} 
                          & DoReFa & \pmb{5.55} & 2.86 & 1.11 & 1.49 \\
                          & PACT   & NC    & \pmb{4.03} & 1.94 & 1.50 \\ 
                          & LSQ    & \pmb{3.73} & 1.78 & 1.49 & 1.36 \\ \bottomrule
\end{tabular}%

\end{table}

\vspace{-0.05in}
\subsection{Evaluation of Natural Corruptions}
\begin{figure}[ht]
  \centering
  \vspace{-0.1in}
  \includegraphics[width=0.99\linewidth]{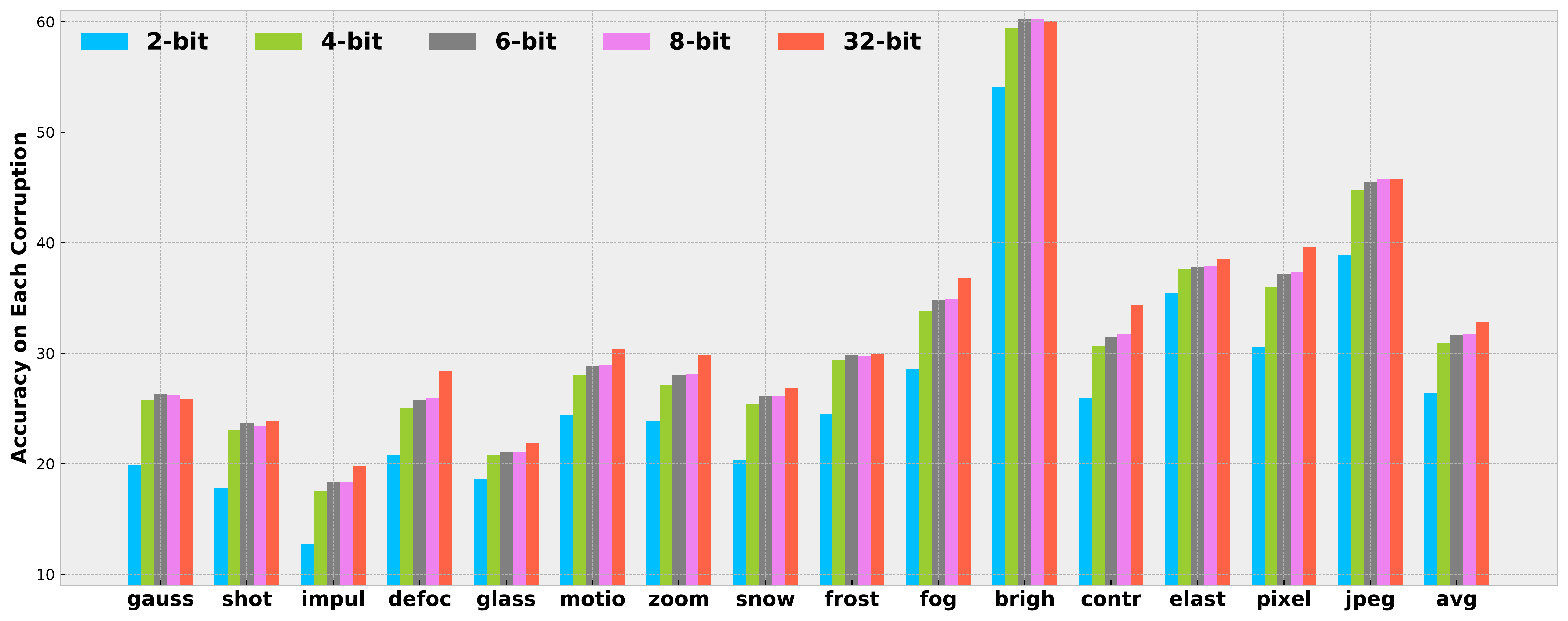}
  \vspace{-0.1in}
  \caption{Accuracy of ResNet18 models quantized by LSQ under each corruption. The `avg' (rightest) is the value of $NR$. }
  \label{fig:resnet18_lsq_avg}
  \vspace{-0.2in}
\end{figure}

\vspace{-0.05in}
We present the Natural Robustness of ResNet18 models in \cref{tab:Corruption-NR} and show the detailed results of LSQ in \cref{fig:resnet18_lsq_avg}. Though performing similar clean accuracy with the 32-bit model, quantized models are more vulnerable under natural corruptions, especially in the 2-bit models. And we can find that \textit{impulse noise} shows the highest impact on the model’s robustness (about 50\% average decrease), while brightness is the least harmful (about 10\% average decrease).

\begin{table}[ht]\small
\centering
\vspace{-0.1in}
\caption{Natural Robustness ($NR\textcolor{red}{\uparrow}$) of ResNet18 models.}
\vspace{-0.1in}
\label{tab:Corruption-NR}
\begin{tabular}{@{}llllll@{}}
\toprule
Model & Method & \multicolumn{1}{c}{2bit} & \multicolumn{1}{c}{4bit} & \multicolumn{1}{c}{6bit} & \multicolumn{1}{c}{8bit} \\ \midrule
\multirow{3}{*}{\makecell{ResNet18 \\ FP: \pmb{32.78}}} 
                          & DoReFa & 23.30 & 30.88 & \pmb{31.79} & 31.70 \\
                          & PACT   & NC     & 30.36 & 31.50 & \pmb{31.69} \\ 
                          & LSQ    & 26.42 & 30.95 & 31.67 & \pmb{31.70} \\ \bottomrule
                          
\end{tabular}%
\vspace{-0.1in}
\end{table}

\vspace{-0.05in}
\subsection{Evaluation of Systematic Noises}
\vspace{-0.05in}
On the systematic noises, we also observe that lower-bit models present less robustness (\ie, lower stability), as shown in \cref{tab:System-SR}. Moreover, we find that among 14 systematic noises, the nearest neighbor interpolation methods in Pillow and OpenCV have the greatest impact on the model performance, which induce nearly a 6\% decrease in performance for the 2-bit models. 
It indicates that maintaining consistency between the deployment and training process is crucial to avoid unnecessary accuracy loss.

\begin{table}[ht]\small
\centering
\vspace{-0.1in}
\caption{Systematic Robustness ($SR\textcolor{green}{\downarrow}$) of ResNet18 models.}
\vspace{-0.1in}
\label{tab:System-SR}
\begin{tabular}{@{}llllll@{}}
\toprule
Model & Method & \multicolumn{1}{c}{2bit} & \multicolumn{1}{c}{4bit} & \multicolumn{1}{c}{6bit} & \multicolumn{1}{c}{8bit} \\ \midrule
\multirow{3}{*}{\makecell{ResNet18 \\ FP: \pmb{0.53}}} 
                          & DoReFa & 2.32 & 1.83 & \pmb{1.69} & 1.95 \\
                          & PACT   &  NC   & 1.85 & \pmb{1.84} & 1.98 \\ 
                          & LSQ    & 2.05 & 1.90 & 1.87 & \pmb{1.82} \\ \bottomrule
                          
\end{tabular}%
\vspace{-0.1in}
\end{table}

\vspace{-0.1in}
\section{Conclusion}
\vspace{-0.05in}
We presents a benchmark for evaluating the robustness of quantized models under various perturbations, including adversarial attacks, natural corruptions, and systematic noises. The benchmark evaluates 4 classical architectures and 3 popular quantization methods with four different bit-widths. The results reveal that lower-bit quantized models have higher adversarial robustness than their floating-point counterparts, but are more vulnerable to natural corruptions (especially impulse noise) and systematic noises (especially the nearest neighbor interpolation). We hope our benchmark will advance the development and deployment of robust quantized models in real-world scenarios.


\noindent\textbf{Acknowledgment} This work was supported by the Academic Excellence Foundation of BUAA for Ph.D. Students.


{\small
\bibliographystyle{abbrv}  
\bibliography{main}
}

\end{document}